\documentclass{article}

    \PassOptionsToPackage{numbers, compress}{natbib}

\usepackage[preprint]{neurips_2025}

\usepackage[utf8]{inputenc} 
\usepackage[T1]{fontenc}    
\usepackage{hyperref}       
\usepackage{url}            
\usepackage{booktabs}       
\usepackage{amsfonts}       
\usepackage{nicefrac}       
\usepackage{microtype}      
\usepackage{xcolor}         

\definecolor{accent}{RGB}{94,129,172}
\definecolor{accent1}{RGB}{129,161,193}
\definecolor{accent2}{RGB}{163,190,140}
\definecolor{accent3}{RGB}{208,135,112}
\definecolor{accent4}{RGB}{143,188,187}
\definecolor{accent5}{RGB}{46,52,64}
\definecolor{accent6}{RGB}{180,142,173}
\definecolor{accent7}{RGB}{143,188,187}
\definecolor{accent8}{RGB}{87,167,115}
\definecolor{accent9}{RGB}{67,146,241}
\definecolor{accent10}{RGB}{141,148,186}
\definecolor{accent11}{RGB}{102,97,123}  
\definecolor{accent12}{RGB}{169,200,218}
\definecolor{accent13}{RGB}{224,204,215}
\definecolor{accent14}{RGB}{194,183,205}
\definecolor{accent15}{RGB}{191,97,106}

\usepackage{subcaption}
\usepackage{lipsum}
\usepackage{graphicx}
\usepackage{ulem} 
\usepackage{color,soul} 

\usepackage{mathtools}
\usepackage{annotate-equations}
\usepackage{placeins}

\newif\ifshowissues

\showissuestrue   

\title{Adapting Lightweight Vision Language Models for Radiological Visual Question Answering}

\author{%
  Aditya Shourya\\
  Department of Advanced Computing Sciences, Maastricht University\\
  \texttt{a.shourya@student.maasstrichtuniversity.nl} \\
    \And
  Michel Dumontier \\
  Institute of Data Science, Maastricht University \\
  Department of Advanced Computing Sciences, Maastricht University\\
  \texttt{michel.dumontier@maastrichtuniversity.nl} \\
  \And
  Chang Sun \\
  Institute of Data Science, Maastricht University \\
  Department of Advanced Computing Sciences, Maastricht University\\
  \texttt{chang.sun@maastrichtuniversity.nl} \\
}

\begin{document}

\maketitle

\begin{abstract}

Recent advancements in vision-language systems have improved the accuracy of Radiological Visual Question Answering (VQA) Models. However, some challenges remain across each stage of model development: limited expert-labeled images hinders data procurement at scale; the intricate and nuanced patterns of radiological images make modeling inherently difficult; and the lack of evaluation evaluation efforts makes it difficult to identify cases where the model might be ill-conditioned. In this study, we fine-tune a lightweight 3B parameter vision-language model for Radiological VQA, demonstrating that small models, when appropriately tuned with curated data, can achieve robust performance across both open- and closed-ended questions. We propose a cost-effective training pipeline from synthetic question-answer pair generation to multi-stage fine-tuning on specialised radiological domain-targeted datasets (e.g., ROCO v2.0, MedPix v2.0). Our results show that despite operating at a fraction of the scale of state-of-the-art models such as LLaVA-Med, our model achieves promising performance given its small parameter size and the limited scale of training data. We introduce a lightweight saliency-based diagnostic tool that enables domain experts to inspect VQA model performance and identify ill-conditioned failure modes through saliency analysis.
Project Link: \url{https://github.com/adishourya/MedM}




\end{abstract}

\section{Introduction} \label{sec_introduction}

Vision-language models (VLMs) have made notable progress in general-domain tasks, such as crop anomaly detection\citep{ultralytics_ai_agriculture_2024} and intelligent video surveillance\citep{verkada_video_alarms}. In the medical and healthcare domain, researchers have recently adapted VLMs to support medical visual question answering (VQA), with promising results from both academic initiatives \citep{photomzroco2023, vansonsbeek2023openendedmedicalvisualquestion} and large-scale efforts \citep{Singhal2023, li2023llavamedtraininglargelanguageandvision}. Alongside improvements in accuracy, recent VLMs have become increasingly accessible to small teams and individual researchers and practitioners to adapt off-the-shelf VLMs to domain-specific tasks through affordable fine-tuning. However, these off-the-shelf VLMs still underperform on medical VQA tasks compared to general-domain VQA due to domain mismatch, limited data availability, and a lack of systematic evaluation and interpretability tools.

Developing robust medical VQA systems poses unique challenges. VLM models are trained on open-web datasets (like \citep{pretrain_pali_changpinyo-etal-2022-may,pretrain_pali_piergiovanni2022pretrainingimagelanguagetransformersopenvocabulary}) that include general-domain data and struggle with the domain shift introduced by complex, multi-modality clinical inputs. Medical VQA tasks require not only visual understanding but also specialized reasoning grounded in clinical knowledge, which general-purpose VLMs typically lack. Moreover, the scarcity of large-scale, high-quality image–question–answer datasets in radiology limits the ability to fine-tune or evaluate these models systematically. In addition, the absence of standardized training pipelines and interpretability tools hampers both model development and clinical validation. Together, these challenges call for lightweight approaches that balance domain adaptation, performance analysis, and interpretability.



We address these challenges by adapting a lightweight VLM - 3B-parameter PaliGemma-mix-448~\citep{beyer2024paligemma} for radiological VQA. Our approach combines a two-stage fine-tuning pipeline with parameter-efficient LoRA \citep{hu2021loralowrankadaptationlarge} adaptation, using a curated mixture of radiology datasets (SLAKE \citep{liu2021slakesemanticallylabeledknowledgeenhanceddataset}, PMC-VQA \citep{xmcmic_pmc_vqa}, ROCO v2.0 \citep{pelka2018roco}, MedPix 2.0 \citep{siragusa2024medpix20comprehensivemultimodal}). In the first stage of fine-tuning, we align the model’s projection head with domain-specific anatomical vocabulary; in stage 2, we fine-tune the full model using enriched instruction-tuning data generated via a LLaMA-8B QA generation pipeline and annealing strategies to amplify high-quality supervision. To evaluate model performance, we introduce a saliency-based diagnostic tool that visualizes attention from image patches to response tokens and vice versa, enabling human experts to identify ill-conditioned outputs. Despite the model’s small size, it achieves competitive accuracy on combined ROCO+MedPix VQA tasks, approaching the performance of much larger models like LLaVA-Med~\citep{li2023llavamedtraininglargelanguageandvision}.

Our \textbf{key contributions} are as follows. First, we reassess model scaling trends in medical VQA by demonstrating that a compact 3B VLM, when appropriately fine-tuned, can achieve competitive performance on radiological VQA tasks, challenging the assumption that only large-scale models are capable of strong clinical reasoning. Second, we propose an end-to-end framework that spans dataset curation, synthetic QA pair generation, annealing-based enrichment, and a two-stage fine-tuning strategy. This pipeline enables medical domain specialization with minimal compute, serving as a practical guide for low-resource medical VLMs. Third, we develop a lightweight, attention-based interpretability tool to visualize cross-modal saliency between image regions and text outputs, supporting expert-driven auditing of model predictions. Finally, we empirically validate our model on both open- and closed-ended radiological QA tasks, highlighting that compact, interpretable models can be viable for domain-specific VQA applications.
\FloatBarrier

\section{Related Work} \label{sec_overview}
Our methodology builds upon recent development in medical VQA and text-based question-answering. Several studies have introduced comprehensive pipelines that span data collection, model training, and rigorous assessment, highlighting the evolving capabilities of a radiological VQA system. We now summarize key contributions from related works that have influenced our approach.


\textbf{MedVInT-T(D,E)} \citep{zhang2024pmcvqavisualinstructiontuning} presents a complete training and evaluation framework for medical VQA. Their approach involves fine-tuning a VLM model on an in-house curated synthetic dataset \citep{xmcmic_pmc_vqa} using GPT-4 \citep{OpenAI_ChatGPT}, which contains multiple-choice style questions to cover a variety of radiological images, and short fill-in-the-blanks style questions with the expectation that the resulting model also develops the capability of answering open-ended queries. The model, fine-tuned on public benchmarks, performs on par with existing radiological VQA systems. Additionally, they manually verify a sample of test set results to make the models robust against the current limitations of popular evaluation frameworks.

  
\textbf{MedPaLM} \citep{Singhal2023}, introduces a comprehensive training and evaluation framework from scratch. They compile HealthSearchQA dataset \citep{healthsearchqa2023} for answering both consumer- and professional-level text-based medical questions by sampling from existing medical QA datasets. They then fine-tune Flan-PaLM \citep{flan_palm} on this dataset, achieving a new state-of-the-art model which is then evaluated by both professionals and laypersons on an extensive set of evaluation axes. Notably, their work exemplifies the design of human evaluation, incorporating assessments from both professionals and laypersons across a broad set of criteria.
  
\textbf{LLaVA-Med} \citep{li2023llavamedtraininglargelanguageandvision} curates PMC-15M dataset by sampling from PubMed Central \citep{pubmedcentral} and prepares synthetically generated multi-turn instruction training data using GPT-4 \citep{OpenAI_ChatGPT}. The study trains the model for only 16 hours on 8xA100 GPUs \citep{nvidia_a100}, achieving state-of-the-art results in radiological visual question answering with a modest 8B LLM \citep{touvron2023llamaopenefficientfoundation}. Their work demonstrates that individual researchers can achieve state-of-the-art performance even with a cost-effective training approach.
  \footnote{PMC-15M \citep{li2023llavamedtraininglargelanguageandvision} remains unavailable to the public at the time of writing.}

\FloatBarrier

\section{Architecture} \label{sec_architecture}
\subsection{Model Design} \label{sec_model_architecture}

\begin{figure}[htbp]
  \centering
  \includegraphics[width=0.9\linewidth]{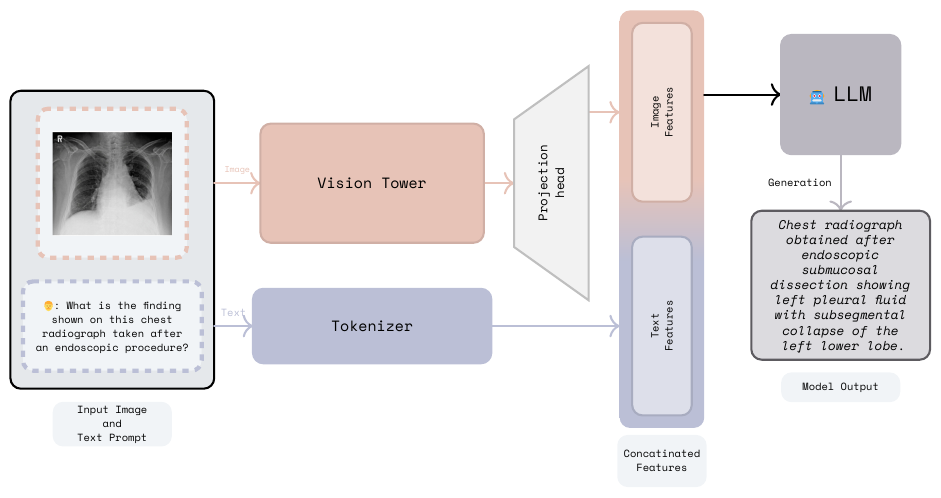}
  \caption{Our and PaliGemma~\cite{beyer2024paligemma} Vision Language Model Architecture}
  \label{fig_model}
\end{figure}

Our vision-language model (VLM) builds on prior work~\citep{beyer2024paligemma, abdin2024phi} and follows a multi-stage training pipeline (Figure~\ref{fig_model}). The training begins with the selection of an off-the-shelf vision-tower and an LLM, each demonstrating strong performance on their respective unimodal tasks, such as large-scale image classification for the vision-tower and natural language understanding and generation for the LLM. These components are then integrated and subjected to multimodal pretraining on a diverse set of tasks such as image captioning \citep{vinyals2015show} and referring expression segmentation \citep{kazemzadeh2014referitgame} to develop a broad understanding of visual concepts in the general domain. During the multimodal pretraining stage of the model, no weights are frozen in time, allowing all parameters to learn during backpropagation.

For domain adaptation such as radiological VQA, we conduct multi-stage fine-tuning on the selected off-the-shelf model using smaller but domain-specific datasets to adapt to specific tasks, mirroring methodologies in~\citep{Singhal2023, li2023llavamedtraininglargelanguageandvision,zhang2024pmcvqavisualinstructiontuning}.
In our study, we employ \textit{PaliGemma-mix-448} \citep{beyer2024paligemma} as our base VLM. This choice is motivated by its transparent pretraining on a diverse and well-curated collection of open-web datasets \citep{pretrain_pali_changpinyo-etal-2022-may, pretrain_pali_piergiovanni2022pretrainingimagelanguagetransformersopenvocabulary, pretrain_pali_sharma-etal-2018-conceptual, pretrain_pali_Srinivasan_2021}, in contrast to models with undisclosed training data \citep{OpenAI_ChatGPT}. This transparency enables a clearer understanding of the model’s zero-shot (base) performance and would make it easier to compare after the base model is fine-tuned. The details about the main components of proposed VLM architecture are described below.

\textbf{Vision Tower} We employ a decoder-only SigLIP transformer~\citep{zhai2023sigmoidlosslanguageimage} as the vision towel in our framework, which contains approximately 400M parameters. pretrained with a sigmoid contrastive loss and comprising ~400M parameters. SigLIP is pretrained using a contrastive learning objective with a sigmoid loss, specifically to handle classification tasks involving a large number of labels where traditional cross-entropy loss becomes less effective \citep{zhai2023sigmoidlosslanguageimage}. The vision tower processes one or multiple input images by applying self-attention across image patches in a non-causal manner, generating image features that are independent of any accompanying text instruction.
  
  
\textbf{Projection Head} A single linear layer aligns the output dimensionality of the vision tower with the token dimension of the language model's vocabulary, which is required for concatenation. While the projection can be implemented using multiple linear layers, the prior ablation study~\citep{beyer2024paligemma} found no significant advantage to have more than one layer. Therefore, we use a single-layer projection in our VLM architecture.

\textbf{Concatenation} The text prefix associated with the image is tokenized ~\citep{kudo2018sentencepiecesimplelanguageindependent}) and concatenated with the projected image features from the vision tower. A special separator token is inserted between the image features and the tokenized text to delineate the two modalities. The resulting sequence is then padded or truncated as needed to match the input length of the language model.

\textbf{LLM} The concatenated image and text features are passed to 2B-\textsc{Gemma} LLM ~\citep{gemmateam2024gemma2improvingopen} as a single input. The model generates the first output token by jointly attending to both the visual features and the tokenized text prefix. Subsequent tokens are produced autoregressively, conditioned on the previously generated tokens along with the original multimodal input.

  

\subsection{Diagnostic Design}


\newcommand{\ScaledDotProductAttnEqn}{
\vspace{2em}
\begin{equation}
\tikzmarknode{Attn}{Attention} = \mathrm{softmax}\left( \frac{\eqnmarkbox[accent8]{Q1}{Q} \eqnmarkbox[accent15]{K1}{K}^\top}{\eqnmarkbox[white]{Scale}{\sqrt{d_k}}} \right) \eqnmarkbox[white]{V1}{V} 
\quad \text{or} \quad 
\mathrm{softmax}\left( \frac{\eqnmarkbox[accent1]{Q2}{Q} \eqnmarkbox[accent15]{K2}{K}^\top}{\eqnmarkbox[white]{Scale}{\sqrt{d_k}}} \right) \eqnmarkbox[white]{V2}{V}
\label{eq_attention}
\end{equation}

\annotate[xshift=-16em,yshift=0.4em]{above}{Q1}{query from selected response token}
\annotate[xshift=-13em,yshift=-1em]{below}{K1}{keys from image patches}

\annotate[xshift=-3em,yshift=0.4em]{above}{Q2}{query from image patch}
\annotate[xshift=2em,yshift=-1em]{below}{K2}{keys from response}
}

To enhance interpretability and validate the clinical relevance of the proposed VQA, we analyze the model’s attention mechanisms, which govern cross-modal interactions between image features and text tokens, inspired by \citep{Mondal_covid_explainability}. 
We develop a diagnostic tool for saliency analysis aimed at aiding practicing radiologists during expert evaluation~\citep{stan2024lvlminterpretinterpretabilitytoollarge}. The interactions between text prefix, image features, and response tokens, which occur exclusively within the attention heads of the LLM, were analyzed with visualizations. Prior to the concatenation layer, there is no interaction between the text prefix and image features. Therefore, the attention heads of the LLM learn to selectively filter and attend to the relevant signals from both modalities to guide the generation process.

Although saliency is not the same as explainability \citep{bertrand2022saliency}, experts can often identify diagnostic indicators, as saliency is fundamentally tied to the learned weights of the model. For a self-attention-based model \citep{vaswani2023attentionneed}, this relation is easy to examine as self-attention operates by aggregating similarity scores between two learned representations for each tokens: queries and keys. These interactions determine how information is distributed across tokens which ultimately guides the generation process. We implemented the following two attention techniques.



\textbf{Saliency via Raw Attention}. Raw attention examines the interactions between queries and keys, which can be interpreted as measuring the affinity or relevance of a token of interest (query) with the rest of the tokens (keys), either within or across modalities. We compute attention weights between queries and keys to localize token-level contributions.

\textbf{Saliency via Rollout Attention}. In self-attention-based models, raw attention weights do not always provide meaningful insights as information propagates through multiple layers, embeddings become increasingly mixed. This is because self-attention does not inherently preserve token identity across layers; rather, it continuously blends representations from multiple input tokens. As a result individual token contributions become obscure, and raw attention weights fail to capture the original token relationship.\citep{chefer_rollout}. We adopt rollout attention  \citep{chefer_rollout,gildenblat_vit_explain}, which recursively aggregates attention weights across layers while accounting for skip connections.

\FloatBarrier

\section{Datasets and Training Recipe} \label{sec_training}




\begin{figure}
    \centering
    \includegraphics[width=0.8\linewidth]{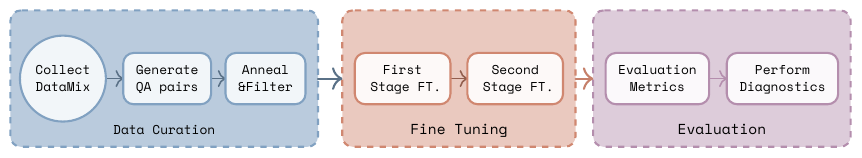}
    \caption{Training Recipe Overview}
    \label{fig:fig_methodology}
\end{figure}

The overall methodology of our training recipe is outlined in Figure~\ref{fig:fig_methodology}. We begin by collecting publicly available radiological datasets and converting them into Visual Question-Answer (VQA) pairs \citep{touvron2023llamaopenefficientfoundation}. The resulting dataset is then enriched and processed to ensure suitability for fine-tuning. Our fine-tuning approach uses a two-stage training strategy: the first stage focuses on learning foundational visual radiological concepts, while the second stage incorporates larger datasets to enhance the model’s rigor and generalization capabilities.

To evaluate model performance, we measure classification accuracy on both open- and closed-ended questions, depending on the dataset composition. For generative responses from open-ended questions, we assess their factuality using GPT-4 \citep{OpenAI_ChatGPT} as an automated judge. We perform ablation studies across different stages of our data curation and finetuning methodology to quantify performance gains. In the absence of a medical expert, the authors of the paper conduct a diagnostic analysis on organ-level cases to identify model limitations.


\subsection{Data Collections}

\begin{figure}[htbp]
    \centering
    \begin{subfigure}[t]{0.24\linewidth}
        \includegraphics[width=\linewidth]{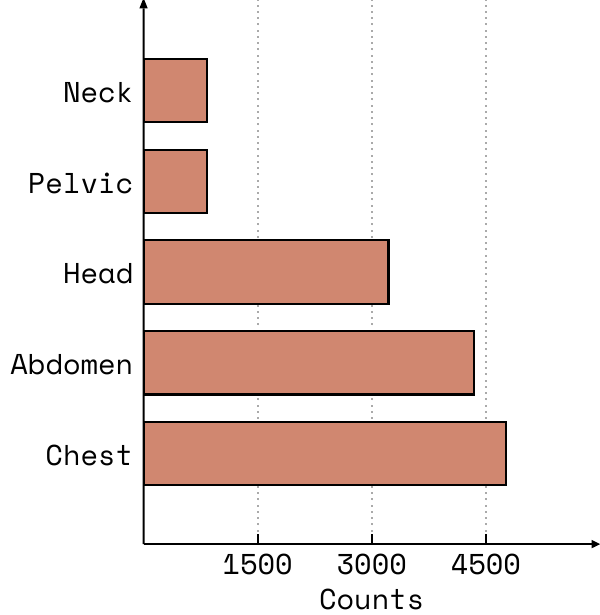}
        \caption{SLAKE: Organ semantic annotations}
        \label{fig_slake_plot}
    \end{subfigure}
    \hfill
    \begin{subfigure}[t]{0.24\linewidth}
        \includegraphics[width=\linewidth]{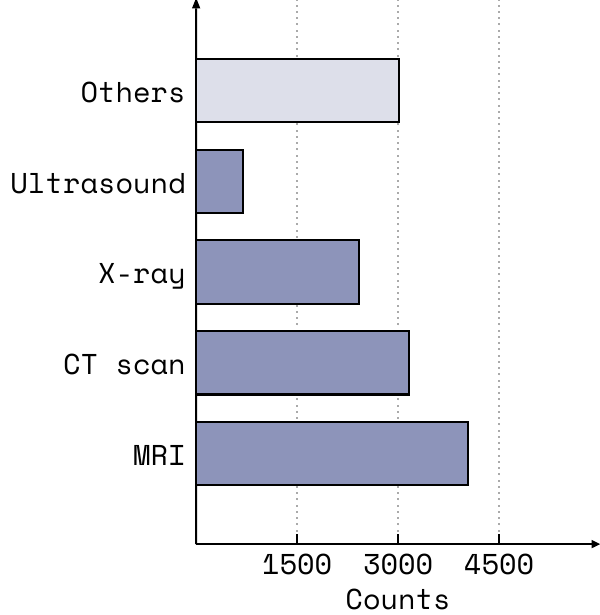}
        \caption{PMC-VQA: Modality distribution}
        \label{fig_pmc_plot}
    \end{subfigure}
    \hfill
    \begin{subfigure}[t]{0.24\linewidth}
        \includegraphics[width=\linewidth]{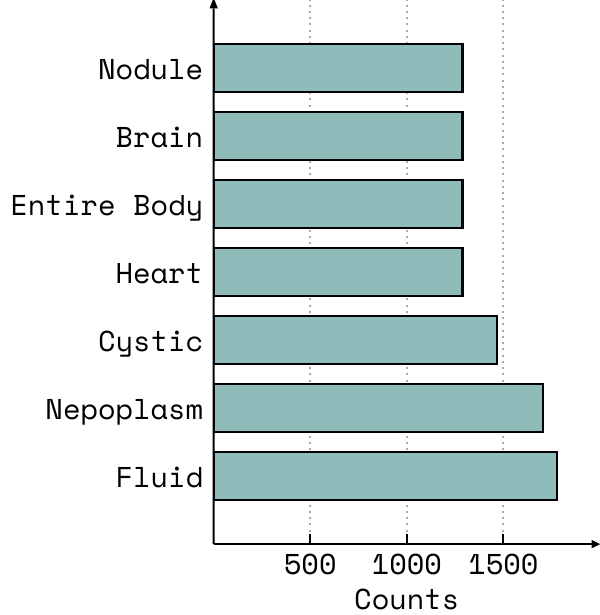}
        \caption{ROCO v2.0: Top7 UMLS concepts}
        \label{fig_roco_plot}
    \end{subfigure}
    \hfill
    \begin{subfigure}[t]{0.24\linewidth}
        \includegraphics[width=\linewidth]{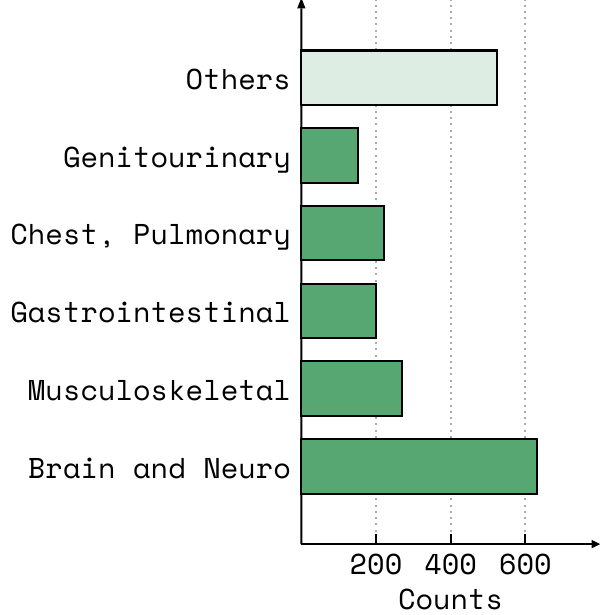}
        \caption{MedPix v2.0: Organ-level distribution}
        \label{fig_medpix_plot}
    \end{subfigure}
    \caption{Distributions across four used datasets.}
    \label{fig:instruct_barchart_grid}
\end{figure}

Fine-tuning VLM requires not only substantial model capacity but also access to large, diverse, and semantically rich datasets. In our work, we combined four datasets that have been de-identified for privacy protection including SLAKE \citep{liu2021slakesemanticallylabeledknowledgeenhanceddataset}, PMC-VQA \citep{xmcmic_pmc_vqa}, ROCOv2 \citep{pelka2018roco}, and MedPix 2.0 \citep{MedPix}. The combination of them spans a wide range of pathodology and radiological modalities (Figure \ref{fig:instruct_barchart_grid}, and concepts for open- and closed-ended questions.

\textbf{SLAKE} contains  $\sim$14{,}000 VQA pairs, annotated by practicing physicians. The dataset covers a wide range of anatomical regions and provides high-quality semantic annotations that are well-suited for evaluating radiological reasoning.

\textbf{PMC-VQA} is derived from PMC-CLIP~\cite{lin2023pmcclip} and includes $\sim$ 227{,}000 QA pairs. The questions are either multiple-choice or short fill-in-the-blank format. Its scale and diversity have been effectively leveraged in training models such as MedVInT-TE and MedVInT-TD. The dataset includes a diverse set of imaging modalities such as CT, MRI, ultrasound, and X-ray.

\textbf{ROCOv2} contains $\sim$79{,}000 image-caption pairs from PubMed Central. Each caption provides a concise ($\sim$ 20 word) description of the radiological images. Due to its breadth and structural consistency, ROCOv2 supports multiple tasks including image captioning, multi-label classification, and VLM pretraining. 

\textbf{MedPix 2.0} includes $\sim$ 12{,}000 curated cases from the MedPix database. Each case contains diagnostic images, detailed case descriptions, and relevant treatment information. The dataset is built using a semi-automated pipeline with manual validation to reduce label noise.



\subsection{QA-Pairs Data Generation} \label{sec_synthetic}


Among our selected datasets, \textit{SLAKE} and \textit{PMC-VQA} natively provide image–QA pairs, while \textit{ROCO v2.0} and \textit{MedPix v2.0} contain image–caption pairs. Fine-tuning on image–QA triplets has been shown to be more effective than image–caption pairs for training VLMs on visual reasoning tasks~\citep{zhu2023chatgptasksblip2answers}. Therefore, inspired by previous work~\cite{brown2020language, zhu2023chatgptasksblip2answers, zhang2024pmcvqavisualinstructiontuning, li2023llavamedtraininglargelanguageandvision}, we synthesize both open- and closed-ended QA pairs from image-caption pairs using LLaMA-8B~\citep{touvron2023llamaopenefficientfoundation}. 
LLaMA-8B was applied for its accessibility, inference efficiency, and reproducibility for other individual researchers. Importantly, its pretraining corpus contains limited medical content, allowing us to isolate and evaluate the performance of general-domain LLMs when applied to specialized medical tasks.

Medical VQA tasks demand not only visual understanding but also clinical reasoning, which general-purpose VLMs often lack. To address this, we prioritize datasets where questions are grounded in patient context and, where possible, linked to supporting medical literature. Figure ~\ref{fig_prompt_qa} and \ref{fig_prompt_lit} in the Appendix show the prompt templates to generate patient case-based and literature-based QA pairs from image-caption pairs. Synthetic QA generation introduces risks such as hallucinations or clinically irrelevant content. To ensure quality, we manually filter out noisy outputs and apply a form of dataset annealing to incrementally refine the corpus toward higher semantic and clinical relevance.



\subsection{Annealing and Filtering}
\begin{figure}
    \centering
    \includegraphics[width=0.8\linewidth]{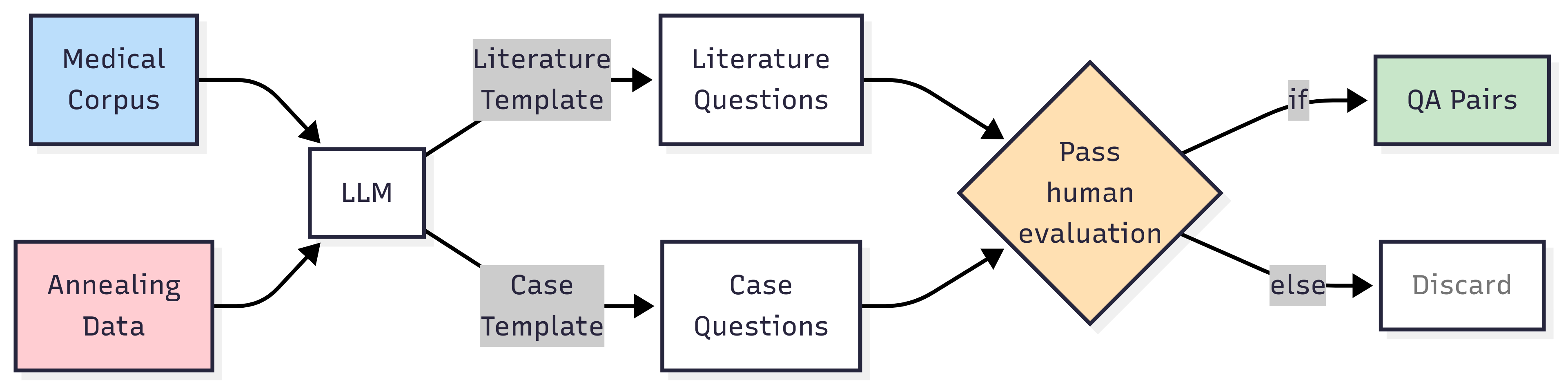}
    \caption{Filtering and curation pipeline.}
    \label{fig_filtering}
\end{figure}

Annealing improves model performance by incrementally incorporating small, high-quality subsets into a larger training set. The objective is to improve the proportion of higher informative examples such as those rich in visual concepts and clinical reasoning within the overall dataset. By doing so, the model will learn more reliable patterns that might otherwise be obscured by lower-quality data.

Evidence for annealing’s effectiveness comes from~\cite{touvron2023llamaopenefficientfoundation}, where LLaMA3-8B showed a 24\% improvement on grade-school-level math questions GSM8K~\cite{cobbe2021gsm8k} and a 6.4\% gain on competition-level math reasoning tasks~\cite{hendrycksmath2021}. Notably, the benefit diminished for larger models (e.g., LLaMA3-405B)~\cite{touvron2023llamaopenefficientfoundation}, suggesting that small or mid-sized models, such as our ~4B parameter VLM, are receptive to annealing.

In our study, we use the high-quality dataset \textit{MedPix v2.0} \cite{siragusa2024medpix20comprehensivemultimodal} as the primary enrichment dataset for annealing \textit{ROCO v2.0}. While \textit{MedPix} is smaller in scale, it provides high-quality radiological case studies and literature references, making it well-suited for improving domain-specific reasoning. A key component of effective annealing is systematic \textit{filtering}, which ensures that only high-quality and domain-relevant data is incorporated into the dataset. Figure \ref{fig_filtering} outlines our data curation stratergy with annealing and filtering. The process begins with a medical corpus, filtered by the pathological relevance. Unlike conventional upsampling strategies that increase the variety of rare cases, our approach focuses on reinforcing the most common pathologies existing in our data mix to improve model generalization.


\FloatBarrier

\subsection{Two-stage Fine-Tuning} \label{sec_scaling}

\newcommand{\FinetuneScalingEquation}{
\vspace{2em}
\begin{equation}
\tikzmarknode{Loss}{\tilde{L}}(X, \eqnmarkbox[green]{Df}{D_f}) = 
\eqnmarkbox[red]{Aterm}{A} 
\cdot \frac{1}{X^{\eqnmarkbox[red]{Alpha}{\alpha}}} 
\cdot \frac{1}{\eqnmarkbox[green]{Df2}{D_f}^{\eqnmarkbox[red]{Beta}{\beta}}}
+ \eqnmarkbox[red]{Eterm}{E}
\end{equation} \label{eq_scaling_law}
}

\newcommand{\ScalingLaw}{
\begin{equation}
\tilde{L}(X, D_f) = 
A \cdot \frac{1}{X^{\alpha}} \cdot \frac{1}{D_f^{\beta}} + E
\label{eq_scaling_law2}
\end{equation} 
}

\textbf{1st Stage}
Off-the-shelf VLMs often exhibit inconsistent performance in recognizing anatomical structures occasionally producing incorrect generation when presented with slight variations in images. This inconsistency highlights the need for alignment between visual features and anatomical vocabulary. To address this, we adopt the SLAKE dataset \cite{liu2021slakesemanticallylabeledknowledgeenhanceddataset} as a foundation for 1st stage fine-tuning. SLAKE offers well-annotated radiological visual concepts, making it particularly suitable for anatomical structure recognition. In this initial phase, we fine-tune only the projection head of the model while keeping all other parameters frozen. We train the projection layer for 5 epochs on SLAKE and use the resulting checkpoint as the initialization point for subsequent model training. Our method aligns with curriculum learning principles, emphasized in \cite{srinivasan2022curriculumlearningdataefficientvisionlanguage}, starting with simpler radiological visual concepts followed by more diverse data.

\paragraph{2nd Stage} Using the checkpoint from the 1st stage as the model weight initialization, we fine-tune the model on larger and more diverse instruction sets - ROCO v2.0 \cite{pelka2018roco}, MedPix 2.0 \cite{MedPix}, and PMC-VQA \cite{PMC_Open_Access}. To perform parameter-efficient fine-tuning, we apply LoRA \cite{hu2021loralowrankadaptationlarge}, a low-rank adaptation method, targeting the attention heads in both the vision tower and the language model. This allows us to significantly reduce computational and storage overhead, deviating from traditional fine-tuning methods that retain all or a large portion of model parameters.

\FloatBarrier

\section{Experiments and Evaluation} \label{sec_evaluation}
\subsection{Experiment Setting} \label{sec_exp_settings}
All experiments including fine-tuning and evaluation were conducted using a single NVIDIA H100 GPU. With adequate allocation, ROCO and MedPix and {Roco + Medpix} datasets were fine-tuned in approximately 3 days each, PMC-VQA takes about 6 days, and SLAKE completes under 5 hours.

\subsection{Fine-Tuning Experiments}

\begin{figure}[t]
  \centering
  \begin{subfigure}[b]{0.42\textwidth}
    \includegraphics[width=\textwidth]{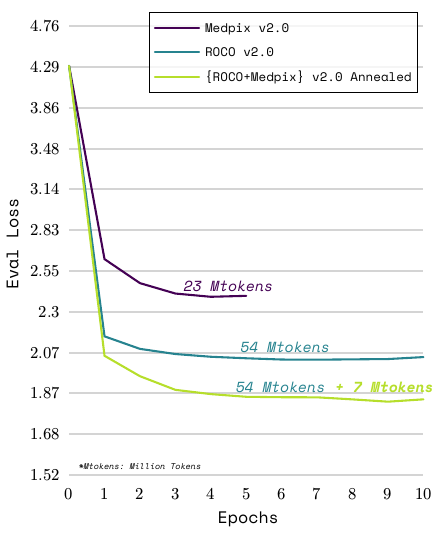}
    \caption{Eval Loss for open ended questions}
    \label{fig_open}
  \end{subfigure}
  \hfill
  \begin{subfigure}[b]{0.42\textwidth}
    \includegraphics[width=\textwidth]{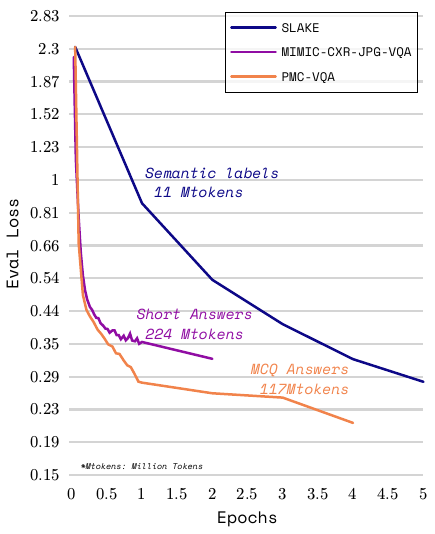}
    \caption{Eval loss for short ended questions}
    \label{fig_closed}
  \end{subfigure}
  \caption{Fine Tuning Evaluation Loss}
  \label{fig:main}
\end{figure}

We first evaluated fine-tuning performance across instruction sets with varying token lengths and question formats. Evaluation loss curves over training epochs are shown in Figures~\ref{fig_open} and~\ref{fig_closed}. For datasets where QA template is open-ended, we observe that the evaluation loss decreases approximately quadratically as the number of tokens in the instruction set increases (Figure~\ref{fig_open}). However, this trend does not hold for datasets with close or short-ended QA templates, where the labels contain fewer tokens as the expected loss after a few training iterations becomes smaller and the loss plateaus earlier (Figure~\ref{fig_closed}).

We further analyzed \textbf{scaling behavior} using the empirical loss model: 
\ScalingLaw
where $\tilde{L}$ is evaluation loss, $X$ is the fine-tuning parameters, $D_f$ is the token size, and $A, \alpha, \beta, E$ are scaling exponents.
Scaling properties for fine-tuning LLMs are highly dependent on task type and data composition \cite{zhang2024when}. Consequently, the optimal fine-tuning strategies and scaling behavior can vary depending on the structure and semantics of the training data. We observed that scaling exponents ($"A"$, $\alpha$,$"\beta"$, $"E"$ in Equation \ref{eq_scaling_law2}) differ depending on the question-answer templates used across datasets. For example, ROCO v2.0 \cite{pelka2018roco} and MedPix 2.0 \cite{MedPix} have open-ended instruction sets with an average label length of around 20 tokens. In this case, task dependence is less observable, and improvements in evaluation loss ($\tilde{L}$) tend to correlate more directly with data size ($D_f$).




In contrast, task dependence becomes more evident in close-ended QA, particularly when different templates are used for QA pairs (Figure~\ref{fig_closed}). While higher data volume generally leads to faster convergence, this trend breaks down when comparing MIMIC-CXR-JPG \cite{johnson2019mimiccxrjpglargepubliclyavailable} and PMC-VQA \cite{xmcmic_pmc_vqa}. Despite its smaller size, PMC-VQA yields greater learning gains in fewer epochs, likely due to the use of multiple-choice templates. These have a lower expected loss ($\tilde{L} = -\ln\left(\frac{1}{4}\right)$
) than open-ended QA tasks, which typically involve more linguistic variation and semantic ambiguity.

These observations suggest that a single scaling law may not generalize across mixed-template datasets. As dataset mixtures grow, especially those combining open- and close-ended QA formats, it becomes increasingly difficult to preserve a consistent ratio of question types.  Since each new addition may introduce variations in this ratio, it becomes challenging to predict the expected evaluation loss as the number of tokens in the instruction set grows. This variability complicates the application of scaling laws in Medical VQA, as the impact of additional training data is not uniform across different datasets and QA templates.


\subsection{VQA Evaluation}

Standard n-gram metrics such as BLEU \cite{papineni2002bleu} and ROUGE \cite{lin-2004-rouge} offer limited insight into factual correctness, particularly in clinical VQA settings~\cite{Singhal2023}. We report these scores in Table~\ref{table_appendix_allstage} in the Appendix, but propose and emphasize more robust evaluation methods below.


\begin{figure}[htb]
    \centering
    \includegraphics[width=0.95\linewidth]{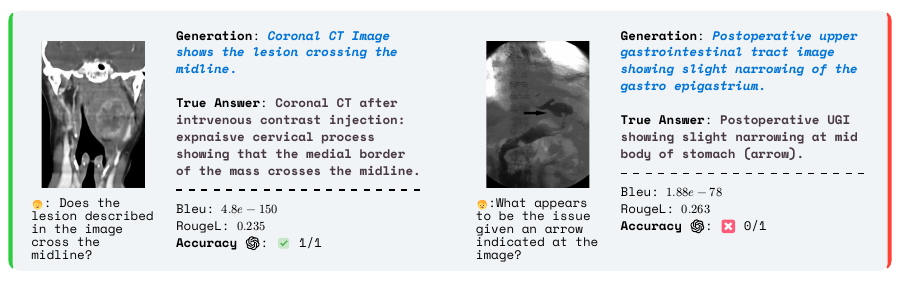}
    \caption{LLM-based Evaluation Examples}
    \label{fig:enter-label}
\end{figure}

\textbf{Closed-ended QA Evaluation}: For multiple-choice question answering (MCQA) such as PMC-VQA \cite{xmcmic_pmc_vqa}, we measure model accuracy across five stochastic generations per test instance. Inspired by \cite{zhang2024pmcvqavisualinstructiontuning}, we define a prediction as \textit{non-robust} if the model produces different answers in three or more out of five inferences. In such cases, we penalize the accuracy by one point to account for uncertainty and instability in the output.

\textbf{Open-Ended QA Evaluation}: For open-ended question that demands clinical reasoning, we employ LLM-based evaluation. We design a prompt template (Figure \ref{fig_prompt_eval} in the Appendix) and use GPT-4.0 \cite{OpenAI_ChatGPT} to judge each generated answer based on factual correctness. Examples are presented in Figure~\ref{fig:enter-label}.

Table~\ref{table_eval1} compares accuracy across four datasets, evaluating the effect of a two-stage fine-tuning approach. Results are reported as the mean accuracy ± standard deviation over five inference runs, with LLaVA-Med serving as a high-capacity baseline. On SLAKE (closed-ended QA), two-stage fine-tuning achieves 79\% accuracy, highlighting strong gains even without large model capacity. For PMC-VQA, ROCO, and the ROCO+MedPix annealing set, two-stage fine-tuning consistently outperforms single-stage fine-tuning, demonstrating its effectiveness across different QA formats. Although the accuracy gains of 2-stage fine-tuning are slight, it accelerated convergence, reducing the number of epochs needed to reach target evaluation loss.
Finally, comparing ROCO to the ROCO+MedPix annealing set shows clear performance gains from annealing, even with small data volumes. These results indicate that modest instruction set annealing offers a cost-effective way to improve generalization and robustness, with potentials for further gains using larger annealing sets.

\begin{table}[ht]
\centering
\small
\resizebox{0.8\linewidth}{!}{
\begin{tabular}{lccc}
\toprule
\textbf{Dataset} & \textbf{w/o Stage 1} & \textbf{with Stage 1} & \textbf{LLaVA-Med} \\
\midrule
SLAKE (Closed)         & --                         & $79.00 \pm 2.75$       & $86.50 \pm 1.60$ \\
PMC-VQA (Short)        & $32.22 \pm 2.23$           & $33.15 \pm 3.15$       & $58.44 \pm 2.53$ \\

ROCO v2.0 (Open)             & $32.25 \pm 3.60$           & $34.00 \pm 3.95$       & $56.56 \pm 3.57$ \\
ROCO + MedPix (Annealing)     & $39.53 \pm 3.22$           & $41.48 \pm 3.90$       & $56.63 \pm 3.22$ \\

\bottomrule
\end{tabular}
}
\caption{Accuracy (\%) with and without Stage 1 fine-tuning across datasets. Results are reported as mean $\pm$ one standard deviation across five inference runs (each on a sample of 200).}
\label{table_eval1}
\end{table}


\subsection{Manual Verification via Saliency Diagnostic}
\begin{figure}[htb]
    \centering
    \includegraphics[width=0.7\linewidth,keepaspectratio=true]{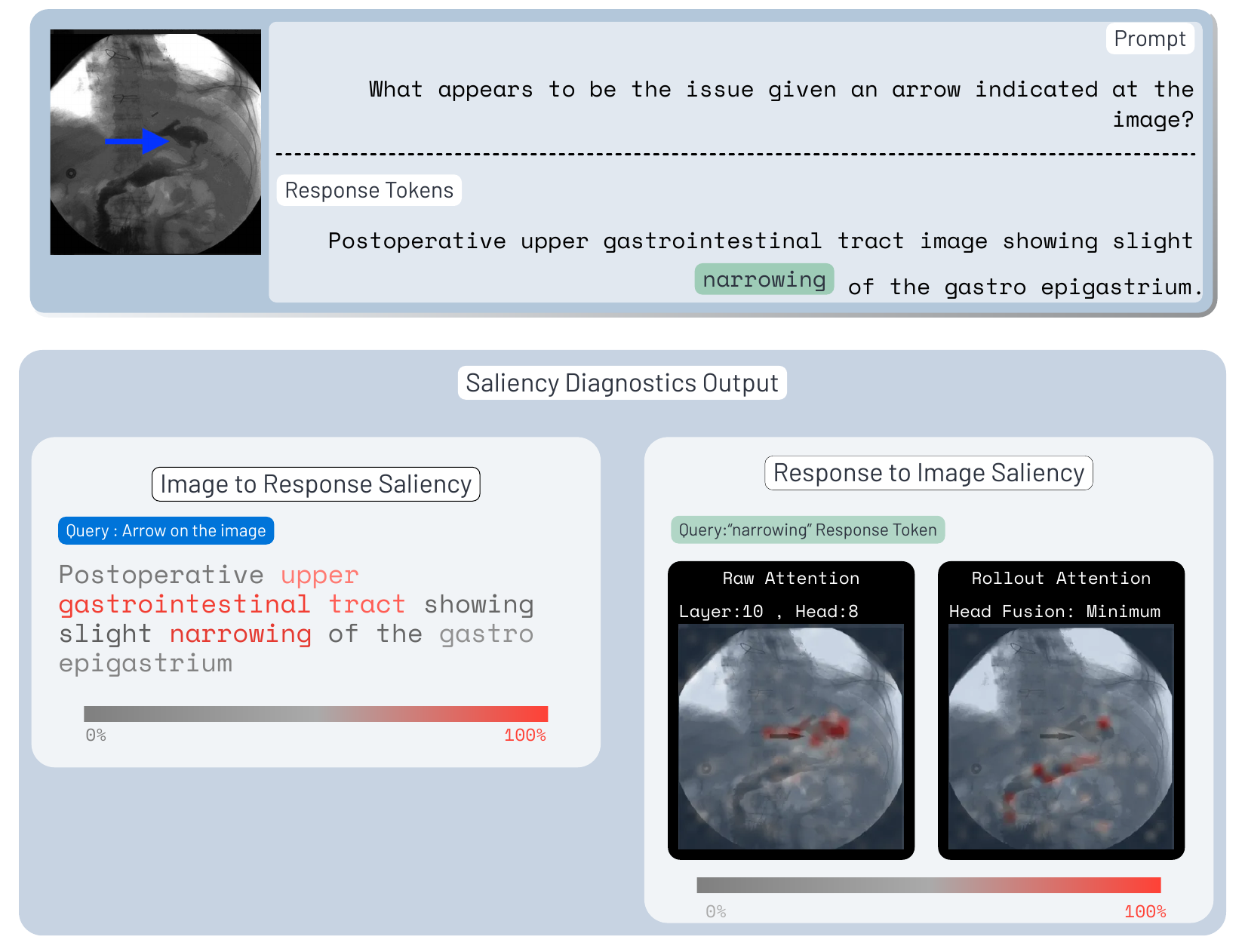}
    \caption{An example illustrates saliency analysis with Raw Attention and Attention Rollout for a patient suffering from a Post-Operative UGI which shows slight narrowing at mid-body of stomach.}
    \label{fig_diagnostic_design}
\end{figure}

Inspired from previous work \citep{zhang2024pmcvqavisualinstructiontuning, Singhal2023}, we conduct manual verification of model-generated responses on test samples, incorporating saliency diagnostic wherever possible for the authors of the study. As discussed above, we applied raw attention and attention rollout methods for saliency analysis. Figure~\ref{fig_diagnostic_design} illustrates an example of these two methods. In the case of response-to-image saliency, we select a response token (e.g., "narrowing") as the query and visualize the average saliency over the input image (used as keys). Conversely, we can also examine image-to-response saliency, where we select a specific image patch (e.g., the blue arrow) as the query and plot the resulting saliency over the response tokens based on their key representations. Compared to raw attention, we found the resulting saliency of attention rollout highlights more abstract features such as the passage of the gastrointestinal tract that are semantically relevant to the given example. More details and examples are presented in Appendix \ref{appendix:saliency}.

Using the saliency tool, we evaluate the factuality of the generated responses from our model against the corresponding ground-truth labels. Furthermore, we report a broad per-class accuracy (Table~\ref{table_2}) at the organ level to highlight variability in model performance, as certain anatomical regions exhibit greater nuance and complexity than others.

\begin{table}[ht]
\centering
\begin{tabular}{lcc}
\toprule
\textbf{Organ-level Pathologies} & \textbf{Accuracy (\%)} \\
\midrule
Chest                          & 15/50 (19/50)\\
Gastrointestinal               & 28/50 (32/50)\\
Musculoskeletal                & 39/50 (41/50)\\
Brain and Neuro                & 14/50 (22/50)\\
\bottomrule
\end{tabular}
\caption{Manual Verification performed over a single inference. (LLava-Med \cite{li2023llavamedtraininglargelanguageandvision} as a baseline)}
\label{table_2}
\end{table}

\FloatBarrier



\section{Conclusion} \label{sec_conclusion}

This study shows that a compact 3B VLM, when fine-tuned with an end-to-end pipeline, can achieve strong performance on radiological VQA tasks. Our framework, including synthetic QA generation, instruction annealing, and two-stage fine-tuning, enables low-resource specialization for medical VLMs. We further introduce a lightweight saliency tool for cross-modal interpretability and validate our approach on both open- and closed-ended QA pairs.

However, several limitations remain. First, our ablation analysis focused primarily on LLM scaling, with a limited investigation into the vision encoder. Second, saliency analysis was conducted without expert involvement, limiting interpretability to broad organ-level patterns. Future work should involve clinicians to evaluate fine-grained anatomical and pathological relevance. Last, our evaluation framework focused only on single-turn QA, whereas real-world clinical workflows involve multi-turn interactions. Expanding the evaluation to multi-turn dialogues would offer a more comprehensive assessment of model reasoning and consistency.
\FloatBarrier

\section{Code and Data Availability} \label{sec_dataavailability}

The code for this project is publicly available at: \url{https://github.com/adishourya/MedM}.
The dataset derived from MedPix v2.0 ~\citep{siragusa2024medpix20comprehensivemultimodal} and used for our annealing experiments can be accessed at:  
\url{https://huggingface.co/datasets/adishourya/MEDPIX-ShortQA}

Synthetically generated question-answer pairs based on the ROCO V2.0 dataset~\citep{pelka2018roco} are available at the following locations:
\begin{itemize}
    \item Training split: \url{https://huggingface.co/datasets/adishourya/ROCO-QA-Train}
    \item Validation and test splits: \url{https://huggingface.co/datasets/adishourya/ROCO-QA}
\end{itemize}


\newpage
\bibliographystyle{unsrtnat}
\bibliography{refs}  

\newpage
\section*{Appendix} \label{sec_appendix}

\FloatBarrier
\appendix

\FloatBarrier

\section{Generating Question Answer Pairs}
\label{appendix_sec_gen_qapairs}


\begin{figure}[h]
    \centering
    \includegraphics[width=0.7\linewidth]{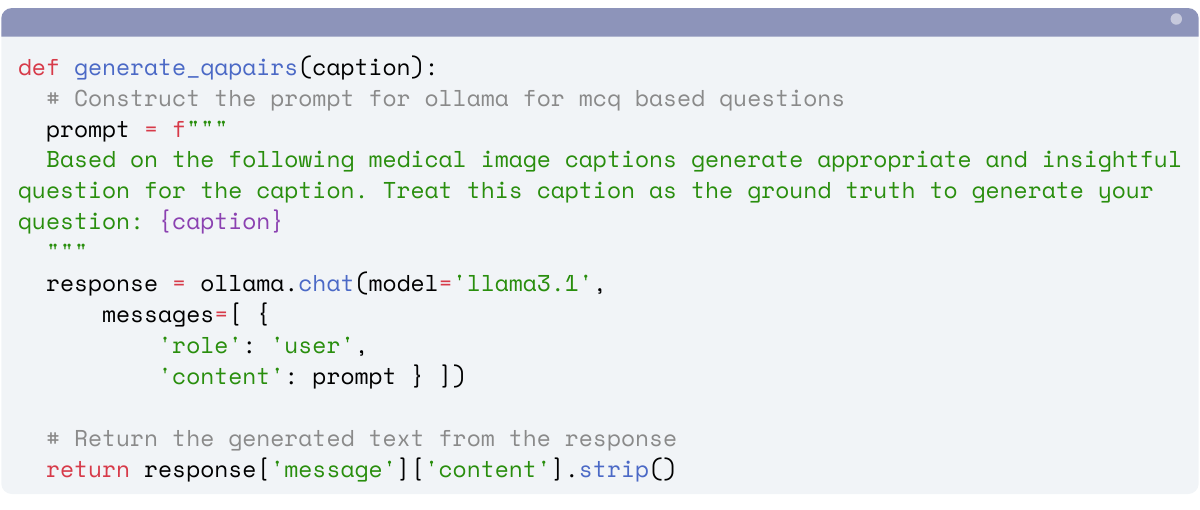}
    \caption{Generate Case Based Questions Prompt}
    \label{fig_prompt_qa}
\end{figure}

\begin{figure}[h]
    \centering
    \includegraphics[width=0.7\linewidth]{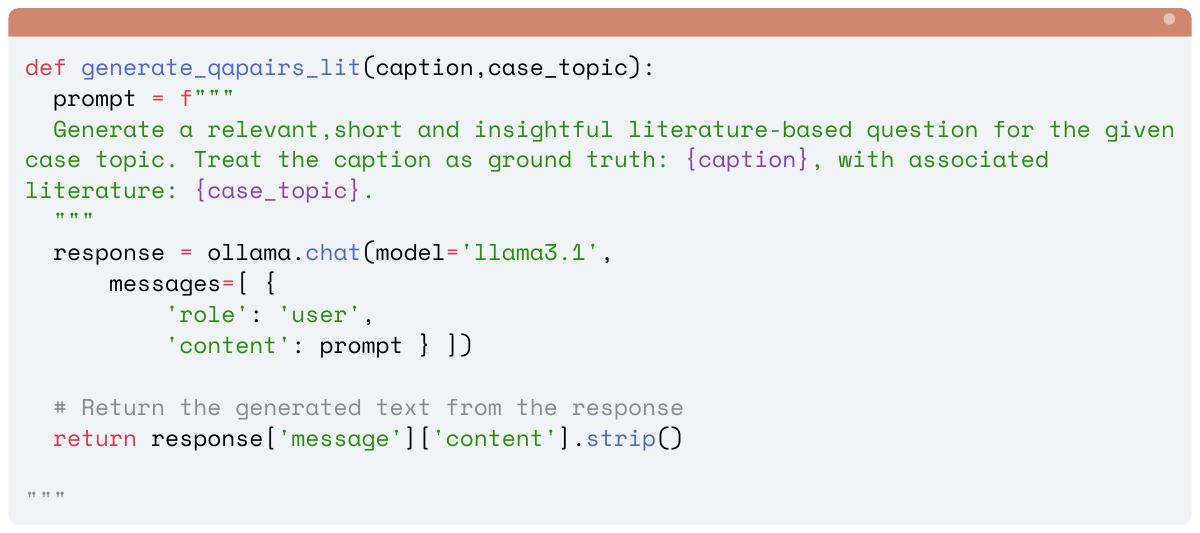}
    \caption{Generate Literature Based Questions Prompt}
    \label{fig_prompt_lit}
\end{figure}

\FloatBarrier
\FloatBarrier
\FloatBarrier


\section{Evaluation and Saliency Diagnostics}
\label{appendix:saliency}
\begin{figure}[h]
    \centering
    \includegraphics[width=0.7\linewidth]{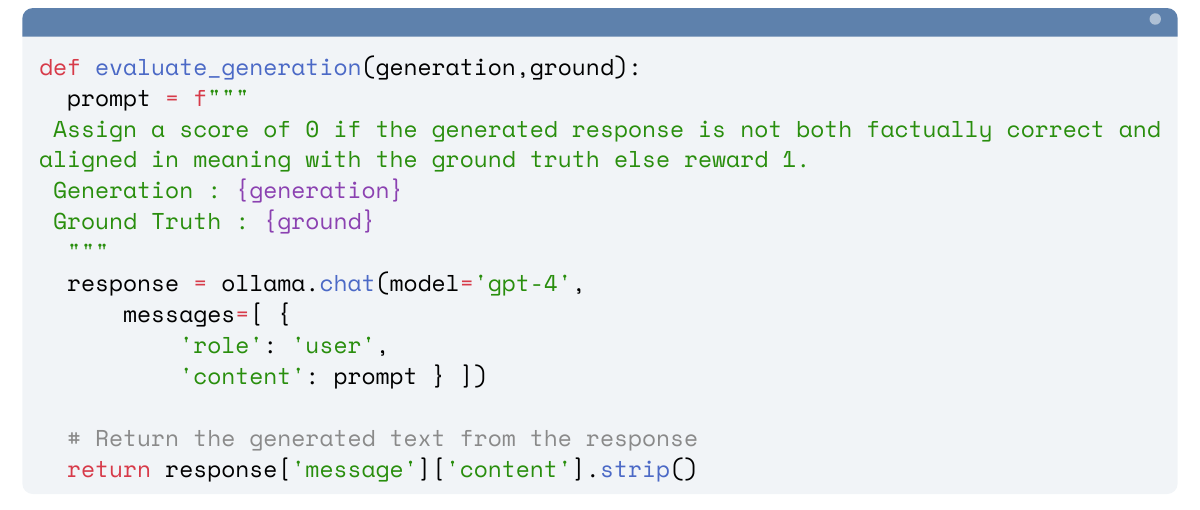}
    \caption{Evaluation prompt for GPT-4 as a judge.}
    \label{fig_prompt_eval}
\end{figure}

\begin{table}[h]
\centering
\begin{tabular}{ll}
\toprule
\textbf{Training Args}       & \textbf{Value} \\
\midrule
learning\_rate               & $1 \times 10^{-5}$ \\
lr\_schedule                 & constant \\
label\_smoothing             & 0.0 \\
weight\_decay                & 0.0 \\
fp16                         & True \\
gradient\_accumulation       & 16 \\
batch\_size                  & 6 \\
\bottomrule
\end{tabular}
\caption{First and second stage training hyperparameters.}
\label{tab:training_hyperparams}
\end{table}

\begin{table}[ht]
\centering
\begin{tabular}{lcccc}
\toprule
\textbf{Metrics} & \textbf{MedPix v2.0} & \textbf{ROCO v2.0} & \textbf{ROCO + MedPix v2.0} & \textbf{PMC-VQA} \\
\midrule
ROUGE-S & $0.311 \pm 0.255$ & $0.325 \pm 0.132$ & $0.334 \pm 0.122$ & -- \\
ROUGE-M & $0.167 \pm 0.082$ & $0.179 \pm 0.124$ & $0.181 \pm 0.124$ & -- \\
ROUGE-L & $0.308 \pm 0.125$ & $0.278 \pm 0.120$ & $0.304 \pm 0.180$ & -- \\
BLEU    & $0.055 \pm 0.111$ & $0.059 \pm 0.090$ & $0.077 \pm 0.065$ & -- \\
\textbf{Accuracy} & $34/200$ \textit{($82/200$)} & $63/200$ \textit{($113/200$)} & $71/200$ \textit{($113/200$)} & $0.3002$ \\
\bottomrule
\end{tabular}
\caption{Results with no first-stage fine-tuning ($\pm$ 1 standard deviation). \textit{LLava-Med}~\cite{li2023llavamedtraininglargelanguageandvision} used as baseline.}
\label{tab:nlp_eval_table_noprestage}
\end{table}

\begin{table}[ht]
\centering
\resizebox{\linewidth}{!}{%
\begin{tabular}{lccccc}
\toprule
\textbf{Metrics} & \textbf{SLAKE (Stage 1)} & \textbf{MedPix v2.0 (Stage 2)} & \textbf{ROCO v2.0 (Stage 2)} & \textbf{ROCO + MedPix v2.0 (Stage 2)} & \textbf{PMC-VQA (Stage 2)} \\
\midrule
ROUGE-S & -- & $0.322 \pm 0.240$ & $0.325 \pm 0.182$ & $0.380 \pm 0.077$ & -- \\
ROUGE-M & -- & $0.165 \pm 0.088$ & $0.196 \pm 0.122$ & $0.219 \pm 0.111$ & -- \\
ROUGE-L & -- & $0.318 \pm 0.125$ & $0.266 \pm 0.080$ & $0.412 \pm 0.109$ & -- \\
BLEU    & -- & $0.024 \pm 0.212$ & $0.008 \pm 0.121$ & $0.430 \pm 0.086$ & -- \\
\textbf{Accuracy} & $0.26$ (\textit{0.59}) & 33/200 (\textit{82/200}) & 68/200 (\textit{113/200}) & 74/200 (\textit{113/200}) & 0.31046 \\
\bottomrule
\end{tabular}
}
\caption{Results $\pm$ 1 standard deviation}
\label{table_appendix_allstage}
\end{table}

\newpage
\subsection{Saliency Diagnostic Examples} \label{sec_appendix_diagnostic_examples}
Note that we examine attention across modalities (Image to Response and Response to Image) as given by Equation \ref{eq_attention}
\ScaledDotProductAttnEqn

\begin{figure}
    \centering
    \includegraphics[width=0.8\linewidth]{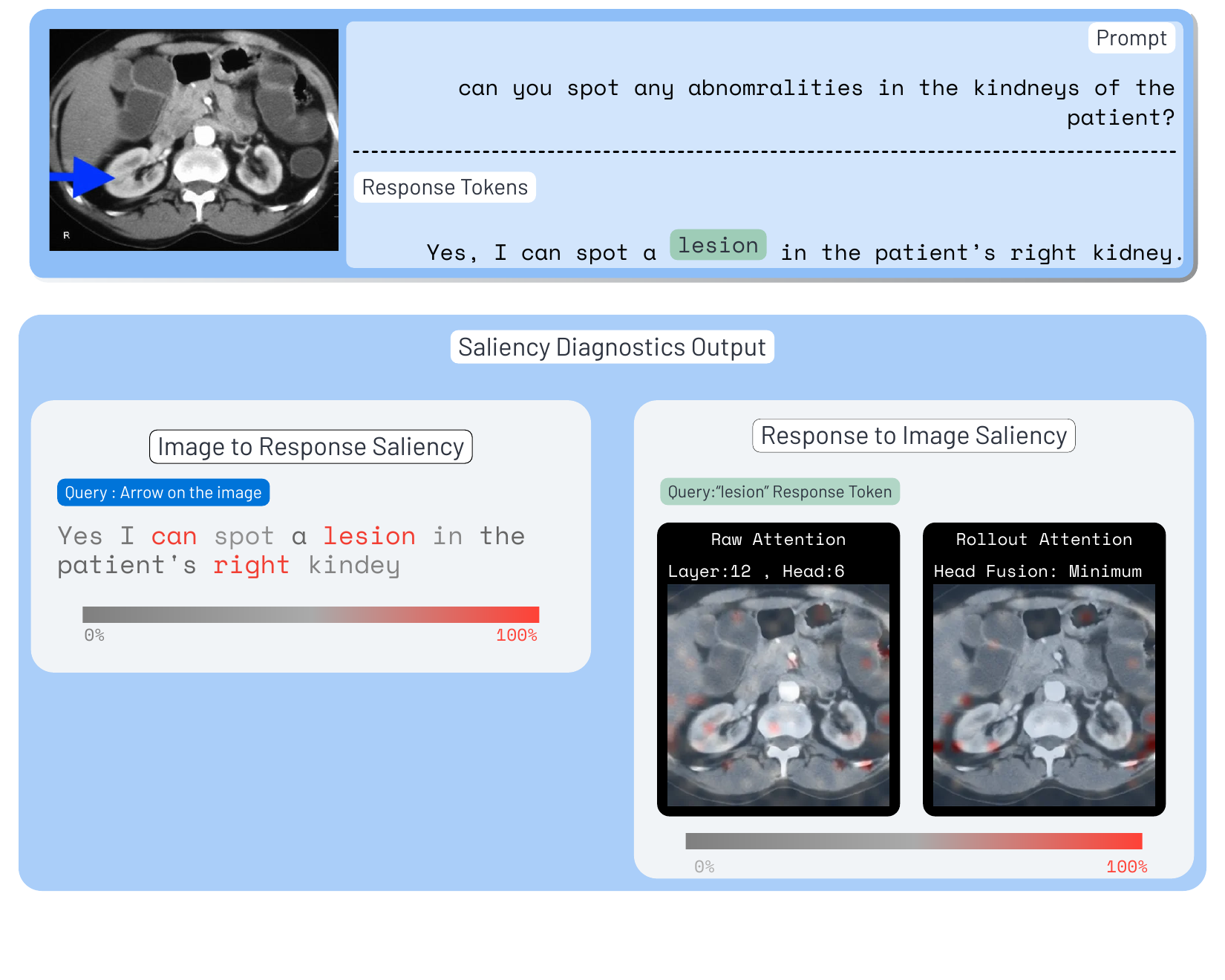}
    \caption{A Patient suffering from lesion on their right kidney [Notice High Rollout Saliency in the right kidney of the patient]}
    \label{fig_lesion_example}
\end{figure}

\begin{figure}
    \centering
    \includegraphics[width=0.8\linewidth]{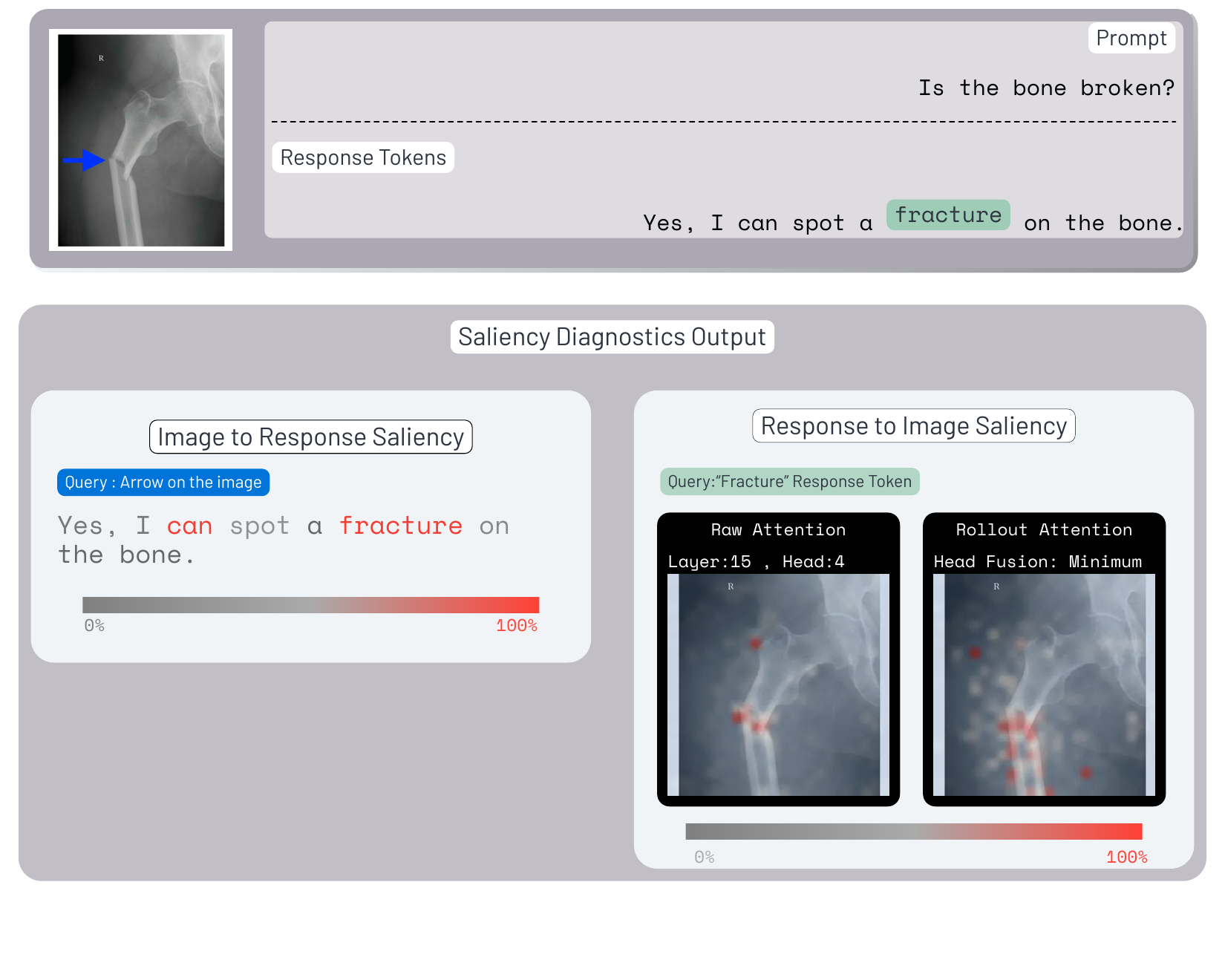}
    \caption{An X-ray of a patient with Bone Fracture[Notice High Saliency on the fractured region]}
    \label{fig_fracture_example}
\end{figure}

\begin{figure}
    \centering
    \includegraphics[width=0.8\linewidth]{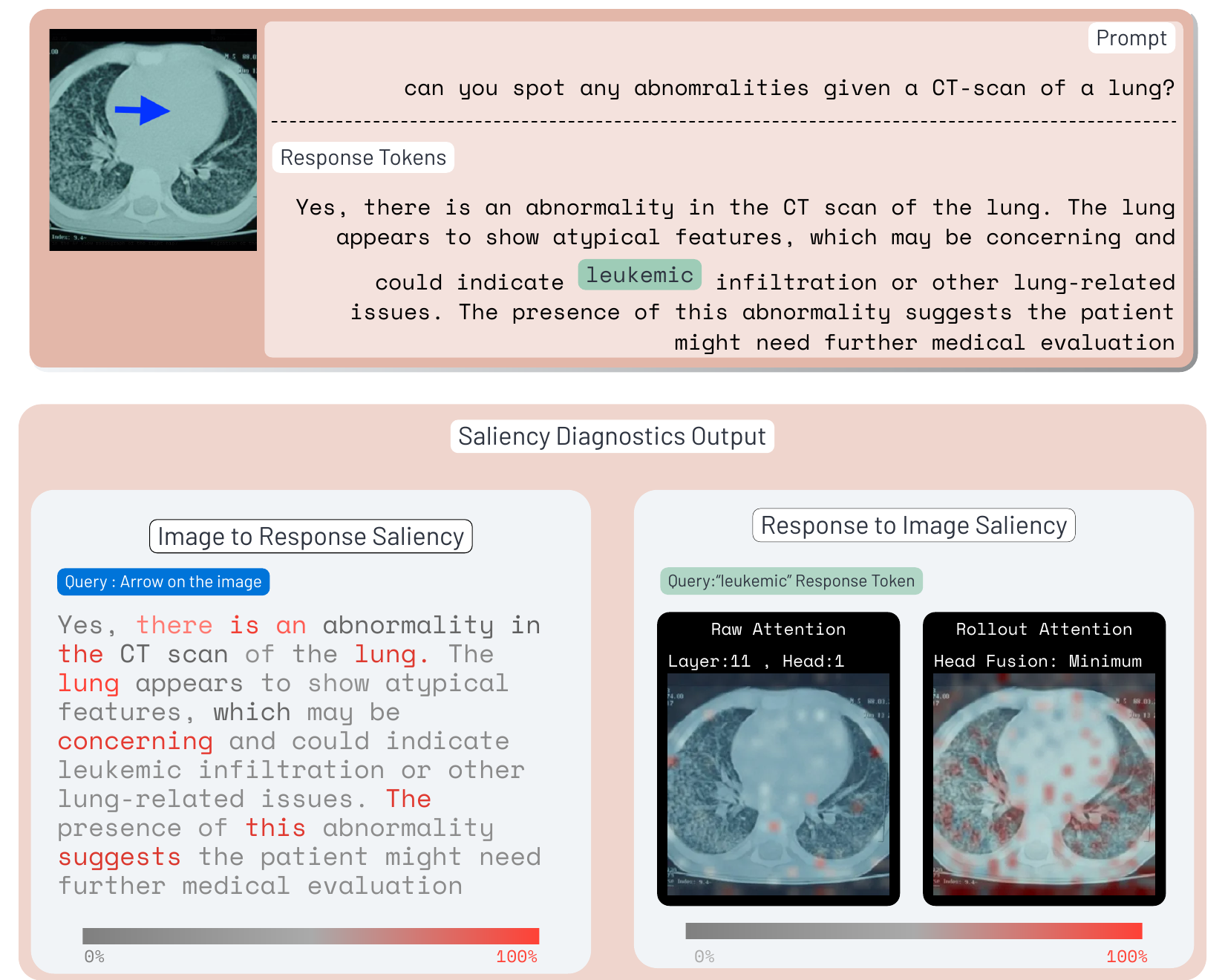}
    \caption{A patient suffering from Leukemia [In such examples the authors of the study refrain from performing saliency diagnostics]}
    \label{fig_lukemia_example}
\end{figure}

\FloatBarrier


\FloatBarrier

\newpage
\newpage

\end{document}